\documentclass[11pt]{amsart}
\usepackage[parfill]{parskip}    
\usepackage{graphicx}
\usepackage{epstopdf}
\newtheorem{theorem}{Theorem}
\newtheorem{lemma}{Lemma}
\newtheorem{corollary}{Corollary}

\newtheorem{notation}{Notation}[]
\newtheorem{hypothesis*}{Hypothesis}[]
\newtheorem{problem}[]{Problem}

\newtheorem{defn}{Definition}
\newtheorem{proposition}{Proposition}
\usepackage{hyperref}
\usepackage{amsthm, amssymb, amsfonts, amsmath, enumerate, xy, mathrsfs}
\usepackage[margin=1in]{geometry}
\input xy
\xyoption{all}
\renewcommand{\tilde}[1]{\widetilde{#1}}

\newcommand{\cyclic}[1]{\langle #1 \rangle}

\title{An Overview of Schema Theory}
\author{David White}
\date{December 30, 2009}

\begin{document}
\maketitle
\begin{abstract}

The purpose of this paper is to give an introduction to the field of Schema Theory written by a mathematician and for mathematicians. In particular, we endeavor to to highlight areas of the field which might be of interest to a mathematician, to point out some related open problems, and to suggest some large-scale projects. Schema theory seeks to give a theoretical justification for the efficacy of the field of genetic algorithms, so readers who have studied genetic algorithms stand to gain the most from this paper. However, nothing beyond basic probability theory is assumed of the reader, and for this reason we write in a fairly informal style.

Because the mathematics behind the theorems in schema theory is relatively elementary, we focus more on the motivation and philosophy. Many of these results have been proven elsewhere, so this paper is designed to serve a primarily expository role. We attempt to cast known results in a new light, which makes the suggested future directions natural. This involves devoting a substantial amount of time to the history of the field. 

We hope that this exposition will entice some mathematicians to do research in this area, that it will serve as a road map for researchers new to the field, and that it will help explain how schema theory developed. Furthermore, we hope that the results collected in this document will serve as a useful reference. Finally, as far as the author knows, the questions raised in the final section are new.
\end{abstract}

\section*{Introduction}

Evolutionary computation began in the 1950s as an attempt to apply the theory of evolution to optimization problems. The goal of an evolutionary program is to begin with a population of candidate solutions for a given problem and then evolve an optimal solution. In order for evolution to occur one needs external pressure, a method by which genetic material is passed to offspring, and time. Genetic algorithms are one way to implement evolutionary computation. A genetic algorithm uses a counter to mimic the passage of time, uses a fitness function to mimic external pressure, and uses reproduction operators to mimic reproduction. 

A genetic algorithm begins with a randomly generated population at time zero. Each iteration of the time counter yields a new \textit{generation}. During any generation, the population is referred to as the \textit{search space}. External pressure is modeled by a fitness function $f$ that assigns (positive) numerical values to candidate solutions. A random process called \textit{selection} determines which solutions survive to the next generation, but solutions with low fitness values have a lower probability of survival. Reproduction is mimicked via operations by which existing solutions produce new solutions. Examples include the crossover operator and the mutation operator. Once a solution evolves whose fitness surpasses some predetermined level, or once a predetermined number of generations pass, the algorithm halts and returns the best solution.

Throughout this paper we will assume that each individual is represented by a string of length $\ell$ taken from the alphabet $\{0,1\}$ (a.k.a. bit-strings). These bits are also referred to as \textit{alleles}, in analogy with evolutionary biology. For an individual $A$ let $A[i]$ denote the bit at position $i$, where the first bit is at $A[0]$. The fitness function is therefore a function from the set of bit-strings to the set of real numbers. The reproduction operations use such strings as input and output. Mutation changes some number of bits at random. Crossover selects two strings uniformly at random, breaks them into pieces, and rearranges the pieces to form two new strings. There are many types of mutation and crossover. An example of mutation is \textit{bit mutation}, which selects a single bit uniformly at random and swaps its value from 0 to 1 or from 1 to 0. An example of crossover is \textit{one-point crossover}, which takes two strings $A$ and $B$, selects a random number $k$ between 1 and $\ell - 2$, and produces new strings $A' = A[0,k]B[k+1,\ell-1]$ and $B'=B[0,k]A[k+1,\ell-1]$. So it simply splits the two strings at index $k$ and then glues their pieces together. 

An example of selection is \textit{fitness proportional selection}, where the probability of selecting an individual $A$ to survive to the next generation is $f(A)/\sum_p f(p)$ where $p$ runs through the population. Another example is \textit{tournament selection}, where a predetermined number of individuals are randomly selected and then divided into a tournament and compared in pairs. A biased coin is flipped, and depending on the outcome the individual which emerges from the pair is either the more fit or less fit of the two. This process continues until only one individual is left. To select $k$ individuals requires $k$ tournaments. In this example, selection pressure can be modified by changing the parameter which represents tournament size or by changing the parameter which represents the coin's bias.

The success of a genetic algorithm often depends on choices made by the programmer. The programmer selects the probability with which crossover or mutation take place in each generation. For example, a selection procedure which is too random may destroy good solutions (i.e. solutions of high fitness) and allow bad solutions to percolate. A selection procedure which is not random enough may encourage the algorithm to evolve an overly naive solution to the optimization problem, i.e. to take the easy way out. Most search heuristics which employ randomness face similar trade-offs, since randomization can always lead to better or worse answers. A more subtle question is to determine how different choices for the fitness function, crossover, and mutation affect the success of the algorithm. To address such questions, and to create more effective genetic algorithms, it is instructive to consider how evolution works from generation to generation rather than as a process spread over thousands of generations.

A first attempt to explain the phenomenon of evolution is the Building Block Hypothesis, which states that crossover combines building blocks (blocks of nearby alleles which endow high fitness) hierarchically into a final solution. Formalizing this statement was the starting point for schema theory. Schema theory was developed in 1975 by John Henry Holland and his work was republished in 1992 (Holland, 1992). Holland realized that when the genetic algorithm evaluates the fitness of a string in the population it is actually evaluating the fitness of many strings in an implicitly parallel fashion. As one of the slowest parts of a genetic algorithm is evaluating fitnesses in the search space, Holland's idea that this could be done in parallel has the potential to speed up genetic algorithms substantially.

\begin{defn}
A \textit{schema} is a string of length $\ell$ from the alphabet $\{0,1,\ast\}$. The $\ast$ is taken to be a symbol for ``don't care" so that one schema corresponds to multiple strings in the search space. The indices where the schema has a 0 or 1 are called the \textit{fixed positions}. The number of fixed positions is the \textit{order} of the schema. When a string matches the schema on the fixed positions it is called an \textit{instance} of the schema, and we say the string \textit{matches} the schema.
\end{defn} 

Thus, a schema is equivalent to a hyperplane in the search space, and we will sometimes use these terms interchangeably. Crossover and mutation can disrupt the schema if they result in a change to one of the fixed positions. A change to one of the other indices results in a different individual in the same schema. We will use the word schema for an individual schema, i.e. a single hyperplane. We will use the word schemata for the plural. 

Schema theory studies how schema percolate in the population and how they relate to the process by which high fitness solutions are evolved over time. The framework of schema theory allows for a definition of a building block and a formal statement of the building block hypothesis:

\begin{defn}
A \textit{building block} is a short, low order schema with above average fitness.
\end{defn}

\begin{hypothesis*}[Building Block Hypothesis] 
Good genetic algorithms combine building blocks to form better solutions.
\end{hypothesis*}

There is a much stronger version of the building block hypothesis which has at times been called the building block hypothesis in the literature, especially literature which is critical of the building block hypothesis. We state it now under a different name to avoid confusion.

\begin{hypothesis*}[Static Building Block Hypothesis] 
Given any schema partition, a genetic algorithm is expected to converge to the class with the best static average fitness. 
\end{hypothesis*}

Readers with a background in biology will recognize that this hypothesis is far too strong. Evolution is far too random of a process for a statement like this to be possible, and indeed empirical studies have demonstrated that the static building block hypothesis is unlikely to be true. See, for instance, (Burjorjee 2008; Syswerda, 1989; Altenberg, 1995; O'Reilly and Oppacher, 1995). 

The building block hypothesis, on the other hand, is sufficiently vague that it has potentially true interpretations. The author believes that the best way to resolve this debate would be to study a genetic algorithm as a dynamical system. Then the building block hypothesis can be given a mathematical interpretation and proven or disproven conclusively.

The controversial nature of the building block hypothesis has led to some unjust criticism of schema theory as a whole. In this paper we hope to convince the reader that schema theory is a useful and interesting study. Towards this end we highlight several open problems and suggested solutions. The author believes that an interested mathematician can make serious headway in this field and in so doing can help to develop the theory behind genetic algorithms and behind evolution in general.

In the next several sections we will summarize the work which has been done in schema theory, beginning in Section \ref{sec:schemaGA}. In this section we also discuss the criticisms of schema theory. The resolution of these criticisms leads into the following sections. The main result of Section \ref{sec:schemaGA} is the Schema Theorem for genetic algorithms. This theorem gives a lower bound on the expect number of instances of a schema. The bound can be turned into an equality, and this is the subject of Section \ref{sec:exact}. The generalizations and strengthenings of the Schema Theorem are phrased in the language of genetic programs and contain the genetic algorithm Schema Theorem as a special case. We choose to begin with the simplest version first so that the reader can become accustomed to the definitions.

In Section \ref{sec:GP} we introduce Genetic Programming, a specialization of genetic algorithms. We discuss how to define schema in this setting, and we give a Schema Theorem regarding the expected number of instances of a given schema. In Section \ref{sec:exact} we state the various Exact Schema Theorems which improve the inequality to an equality. These sections form the heart of the paper, and afterwards the reader could safely skip to the final section, which includes a list of open problems.

The development of schema theory saw a large number of definitions and varied approaches, some of which are cataloged in Section \ref{sec:GP}. Most approaches faced stiff criticism and needed to be tweaked over subsequent years. For this reason it may be hard for an outsider to determine the current status of the field. This paper is meant to provide a unified treatment of the development and current state of the field, to clearly state the definitions which appear to have been most successful, to list a number of applications of the field which have been spread across several papers, and to highlight directions for future work.

Readers interested in developing more advanced applications of probability theory to schema theory may be interested in Sections \ref{sec:expectation}, \ref{sec:conditional}, and \ref{sec:markov}, as these will help the reader find the boundaries of the field. In Section \ref{sec:applications} we discuss applications of the schema theory developed in the earlier sections. Finally, a large list of open problems can be found in Section \ref{sec:future}, and the author hopes this will lead to more development of this beautiful field.

\section{The Schema Theorem for Genetic Algorithms} \label{sec:schemaGA}

There are a number of important functions related to a schema, and we list here the standard notation which will be used throughout the paper:

\begin{notation} \label{notation} For a fixed schema $H$:
\begin{itemize}
\item The order $o(H)$ is the number of fixed positions in the string.

\item The defining length $d(H)$ is the distance between the first and last fixed positions, i.e. the number of places where crossover can disrupt the schema.

\item The fragility of the schema is the number $\frac{d(H)}{\ell-1}$, i.e. the proportion of places where the schema can be disrupted.

\item The fitness $f(H,t)$ of a schema is the average fitness of all strings in the population matching schema $H$ at generation $t$. Let $\overline{f}(t)$ denote the average of the $f(-,t)$ values. Let $\overline{f}$ denote the average fitness of the whole population at time $t$.
\end{itemize}
\end{notation}

Compact schema are those with small defining length. These are the schema which are less likely to be disrupted by crossover. The maximum number of compact schema is $2^{\ell-o(H)}$. Note that not every subset of the search space can be described as a schema since there are $3^\ell$ possible schemata but $2^{2^\ell}$ possible subsets of the search space. A population of $n$ strings has instances of between $2^\ell$ and $n\cdot 2^\ell$ different schemata. A string of length $\ell$ can match up to $2^\ell$ schemata. 

In the language of schema theory, Holland's implicit parallelism is a statement about the number of hyperplanes sampled by a single string. In this way, evaluating the fitness of a string gives information on all the schemata which the string matches. 

\begin{proposition}[Implicit Parallelism]
A population of size $n$ can process $O(n^3)$ schemata per generation, i.e. these schemata will not be disrupted by crossover or mutation.
\end{proposition}

This result holds when $64 \leq n \leq 2^{20}$ and $\ell \geq 64$. Let $\phi$ be the number of instances of $H$ in the population needed to say we've ``sampled" $H$. This is a parameter which a statistician would set based on the certainty level desired in the statistical analysis. Let $\theta$ be the highest order of hyperplane which is represented by $H$ in the population. Then $\theta = log_2(n/\phi)$ and some basic combinatorics shows that the number of schemata of order $\theta$ in the population is $2^\theta \cdot {\ell \choose \theta} \geq n^3$.

These types of consideration demonstrate that the language of schema theory allows for the methods of statistics to be used in the analysis of genetic algorithms. We can see that it is valuable to be able to count instances of a given schema in the population. 

Assume now that the genetic algorithm is using one-point crossover and bit mutation with probability $p_c$ and $p_m$ respectively (i.e. these are the probabilities with which the operator is used in a given generation). The first attempt to count the number of instances of a fixed schema in a genetic program led to the Schema Theorem. The version stated below is slightly more general than Holland's original version because it factors in mutation fully rather than assuming $p_m$ is approximately zero.

\begin{theorem}[Schema Theorem]
Let $m(H,t)$ is the number of instances of $H$ in the population in generation $t$, and let $f$ denote the fitness as in Notation \ref{notation}. Then

$$E(m(H,t+1)) \geq \frac{f(H,t)}{\overline{f}(t)}\cdot m(H,t)\left(1-p_c\frac{d(H)}{\ell-1}\right)(1-p_m)^{o(H)}$$
\end{theorem}

The proof is an exercise in elementary probability. The result is an inequality because Holland only factored in the destructive effects of crossover and assumed that every time crossover occurred within the defining length it destroyed the schema. In Section \ref{sec:exact} we will state the Exact Schema Theorem, which obtains an equality by factoring constructive effects as well.

The Schema Theorem can be interpreted to mean that building blocks will have exponentially many instances evaluated. This does not directly support the building block hypothesis because there is no mention of crossover or how these building blocks are used in the creation of a final solution.

There are many applications of schema theory to both the theoretical study of genetic algorithm and the practical use of genetic algorithm. One of the first was Holland's \textit{principle of minimal alphabets}, which gave an argument based on the Schema Theorem for why a binary encoding is optimal. Unfortunately, this was an informal argument, and it led to some criticism of the emerging field of schema theory as a whole. 

\subsection{History of Schema Theory}

After Holland's work was republished in 1992, there was a strong interest in schema theory from the computer science community. The principle of minimal alphabets was hotly debated and detractors produced convincing arguments for why larger alphabets are optimal for certain problems. Various computer scientists created versions of the Schema Theorem which held with different crossover, mutation, and selection. Riccardo Poli and others took up the challenge of generalizing the Schema Theorem for genetic programming and this brought the debate over the usefulness of the Schema Theorem to the fore. The following quote nicely sums up the world of schema theory as it existed in the mid 1990s:

``Some researchers have overinterpreted these approximate schema models, leading to the formulation of hypotheses on how genetic algorithms work, first highly celebrated and later disproved or shown to be only first order approximations. Other researchers have just incorrectly interpreted such models, while still others have understood and correctly criticized them for their limitations. All this has led to the perception that schema theories are intrinsically inferior to other models and that they are basically only a thing of the past to be either criticized to death or just be swept under the carpet." (Poli, 2001a)

Poli first published on the Schema Theorem in 1997 (Poli and Langdon, 1997) and subsequently published numerous papers on schema theory every year until 2001. During this time comparatively few others were publishing in the field. Poli successfully defended the Schema Theorem against every attack this author has found, and he deserves a great deal of credit for the existence of the field today. Poli last published in schema theory in 2004, and that work is summarized in Section \ref{sec:markov}. 

\subsection{Criticisms of Schema Theory}

When Poli began, schema theory was a fairly unpopular field. The most common criticisms of the Schema Theorem (e.g. Whitley, 1993) follow. First, the Schema Theorem is only useful to look one generation into the future rather than many due to its reliance on the expectation operator and the fact that it is an inequality. Too much information is lost by not considering the constructive effects of crossover. Second, because selection becomes biased as the population evolves, the observed fitness of schema changes radically. Thus, the average fitness of a schema is only relevant in the first few generations of a run. Thus, the Schema Theorem cannot be used recursively. Third, schema have no useful applications.

One response to the first criticism is to let the population size approach infinity so that we can remove the expectation operator via the Law of Large Numbers. Although arbitrarily large population sizes are impractical, letting $n\to \infty$ has become a common response in theoretical computer science and especially in the asymptotic analysis of algorithms. If Whitley's criticism of schema theory is accepted then it extends to a criticism of much of modern computer science. A different approach is to use Chebychev's inequality from probability theory to remove the need for the expectation operator. For more on this see Section \ref{sec:expectation}. However, as is the case with classical probability theory, using Chebychev's inequality involves a loss of information. For this reason, the author prefers the other responses to the first criticism.

The second criticism was resolved in (Poli, 2000a) and (Poli, 2003b) in the more general setting of genetic programming. A particularly strong version of the Schema Theorem is the Recursive Conditional Schema Theorem, and this resolves the second criticism very completely. The goal of reaching this theorem serves as motivation for our introduction of this language in Section \ref{sec:GP}. This work requires the use of more tools from probability theory than the Schema Theorem above. At several points during the development of schema theory, new ideas from probability theory propelled the field forward. Examples can be found in Sections \ref{sec:exact}, \ref{sec:conditional}, and \ref{sec:markov}. These developments required studying genetic programs microscopically rather than macroscopically, the use of conditional probability, and the use of methods from the theory of Markov chains.  For future development in this vein, the author suggests looking to the theory of dynamical systems for more tools which may be used here.

The third criticism can again be taken as a criticism of all of theoretical computer science. As is often the case, applications of schema theory did eventually arise, and these are cataloged in Section \ref{sec:applications}. One of the nicest applications of the ideas in this section came not from the schema theory of genetic algorithms, but rather from the version for genetic programs. We address this version in the next section.

\section{Schema Theory for Genetic Programming} \label{sec:GP}

A genetic program is a special type of genetic algorithm in which each individual in the population is now a computer program. An optimal solution is therefore an optimal algorithm for completing a predetermined task, and so the theory of genetic programming falls under the umbrella of machine learning. The types of considerations which go into genetic programming are related to those of genetic algorithms, but more complicated because individuals are now more naturally represented as parse trees rather than strings. For this reason, crossover and mutation are now operations on trees and there are numerous ways to define these operations, each with their own advantages and disadvantages. Unlike genetic algorithms, the size of individuals in genetic programming cannot be restricted. Trees are therefore allowed to grow arbitrarily large as the number of generations grows, and this phenomenon (called \textit{bloat}) is common. The interested reader is encouraged to consult (Koza, 1992) for further details.

Because genetic programming is a generalization of genetic algorithm, all schema theorems for the former apply to the latter. Indeed, in (Poli 2001a) it was shown that these theorems in fact apply to genetic algorithm with variable length bit-string representation. Much of Poli's work is phrased in terms of genetic programs, and that is our motivation for introducing them.

The notion of a schema for genetic programming was studied by several authors in the 1990s, and numerous non-equivalent definitions arose (Koza, 1992; Altenberg, 1994; O'Reilly and Oppacher, 1994; Whigham, 1995). These papers all used representations for genetic programs which did not include positional information and which thus allowed numerous instances of a schema in one program. By way of analogy, the genetic algorithm schema could be defined as a set of pairs $\{(c_1, i_1), (c_2, i_2), \dots\}$ where the $c_j$ are blocks of bit-strings with no breaks and the $i_j$ are positions for the $c_j$. Thus, the position of a schema instance is implicit in the genetic algorithm situation. Removing this positional information can lead to the number of strings belonging to a given schema not being equal to the number of instantiations of the schema in the population. Without positional information, counting the number of schema and dealing with even the simplest quantities in the genetic algorithm schema theorem becomes much harder. 

In (Poli and Langdon, 1997), the authors considered all existing definitions of genetic program schema and then created their own. Their definition simplified the calculations and allowed for the first non-trivial genetic programming schema theorem. This is why Poli and Langdon created their own definition. As motivation for Poli-Langdon schema, we will record the historical definitions here.

\subsection{Koza's Definition}

Koza defined a schema as a subspace of trees containing a predefined set of subtrees. O'Reilly and Oppacher's definition was a formalization of Koza's with ``tree fragments" which are trees with at least one leaf that is a ``don't care" symbol, represented by $\#$. These symbols could be matched by any subtree. With this definition, one schema can appear multiple times in the same program. It's clear how to define the order and defining length for one instance, but because it depends on which instance is chosen these quantities are not well-defined for the schema. Still, O'Reilly and Oppacher were able to craft a schema theorem for genetic programming using this definition when fitness proportional selection is used. Before stating this theorem, we record our notation for the section

\begin{notation}\label{GPnotation}

For a fixed genetic programming schema $H$:

\begin{enumerate}
\item $m(H,t)$ is the number of instances of $H$ at generation $t$.

\item $P_{d_c}(H,h,t)$ is the probability of disruption due to crossover of schema $H$ in program $h$ in generation $t$. Taking the average over all $h$ yields $\overline{P}_{d_c}(H,t)$. Analogously we obtain $P_{d_m}$ for mutation.

\item The notation for $f$ and $\overline{f}$ is the same as in Notation \ref{notation}.

\end{enumerate}

\end{notation}

We now state O'Reilly and Oppacher's theorem, using the notation above

$$E(m(H,t+1))\geq m(H,t)\frac{f(H,t)}{\overline{f}(t)} \left(1-p_c\cdot \max_{h\in Pop(t)}P_d(H,h,t)\right)$$

O'Reilly and Oppacher state the \textit{genetic programming building block hypothesis} as: ``genetic programs combine building blocks, the low order, compact highly fit partial solutions of past samplings, to compose individuals which, over generations, improve in fitness."

They observe that a schema instance can change in a generation even when the schema instance is not disrupted. They use this to question the existence of building blocks in genetic programming. They also attack the usefulness of schema theory with the same arguments used in (Whitley, 1993). They gave two criticisms of schema theory: first, that it could never be correctly applied to genetic programming because genetic programs exhibit time dependent behavior similar to a stochastic processes. Second, that genetic programming exerts no control over program size but program size matters quite a bit to performance. These issues were not resolved until (Poli, 2000a) and (Poli, 2003b).

\subsection{Whigham's Definition}

Whigham's paper addresses context free grammars, which are generalizations of genetic programs. Context free grammars may be thought of as derivation trees which explicitly tell how to apply rewrite rules coming from a pre-defined grammar $\Sigma$ to a starting symbol $S$. The result of this application is the program. The derivation tree has internal nodes which may be thought of as rewrite rules and terminals which may be thought of as the functions and terminals used in the program. Whigham's schema are partial derivation trees rooted in some non-terminal node. Formally, a schema is a derivation tree $x \Rightarrow \alpha$ where $x$ is in the set of non-terminals $N$ and $\alpha \in \{N\cup \Sigma\}^\ast$. Thus, one schema represents all programs which can be obtained by adding rules to the leaves of the schema until only terminal symbols are present. As with O'Reilly and Koza, one schema can occur multiple times in the derivation tree of the same program, but Whigham also is able to obtain a schema theorem in the presence of fitness proportionate reproduction. We state this theorem, again using Notation \ref{GPnotation} and our usual notation regarding probabilities $p_c$ and $p_m$ of crossover and mutation:

$$E(m(H, t + 1)) \geq m(H,t)\frac{f(H, t)}{\overline{f}(t)} \cdot ((1-p_m \overline{P}_{d_m} (H, t))(1 - p_c \overline{P}_{d_c} (H, t)))$$

The definitions for mutation are similar. Whigham also has an equivalent statement of the theorem using only probabilistic quantities. He uses his theorem as a possible explanation for the well-known problem of bloat in genetic programming, but does not rigorously show anything to this effect. He does show, however, that his notion of schema and his schema theorem applies to fixed-length binary genetic algorithms under one-point crossover and to genetic programming under standard crossover.

\subsection{Altenberg's Definition}

Altenberg focused on emergent phenomena, meaning ``behaviors that are not described in the specification of the system, but which become evident in its dynamics." An example of such behavior is bloat. Altenberg defines soft-brood selection in order to avoid producing offspring of low fitness. This selection method is very similar to tournament selection but there is no probability that the tournament winner will be selected randomly and there is a probability that no recombination will occur. Let $x_i$ be the frequency of program $i$ in the population, $P$ be the space of programs, $S$ be the space of subexpressions which recombination can obtain from programs in $P$ , $C(s\gets k)$ be the probability that recombination obtains $s$ from program $k$, and $P (i\gets j, s)$ be the probability that recombination on subexpression $s$ and program $j$ produces program $i$. Then without soft-brood selection,

$$x_{i+1} = (1-\alpha)\frac{f(i)}{\overline{f}(t)}x_i + \alpha \sum_{j,k\in P}\frac{f(j)f(k)}{\overline{f}(t)^2}x_jx_k \sum_{s\in S}P(i\gets j,s)C(s\gets k)$$

Altenberg uses this to create a schema theorem for genetic programming in the situation of an infinite population. Suppose $v_s$ is the marginal fitness of subexpression $s$ obtained from averaging weighted sums of $C(s\gets i)$. Suppose $\overline{u}_s(t)$ is the weighted sum $\sum_{i\in P}C(s\gets i)x_i$ in generation $t$. Then

$\overline{u}_s(t+1) \geq (1-\alpha)v_s\overline{u}_s(t)/\overline{f}(t)$

Altenberg then factors in soft-brood selection and analyzes how it changes the situation. His schemata are the subexpressions $s$ and as with the definitions above, each schema lacks positional data and so can appear multiple times in a given program $i$. One of Altenberg's models is an exact microscopic model (i.e. the constructive effects of crossover are taken into account), but it fails to be a theorem about schemata as sets. Such a model did not exist in Poli's work until 2001.

\subsection{Rosca's Definition}

Justinian Rosca (Rosca 1997) published a new definition of genetic program schema at the same time as Poli and Langdon. Rosca's definition of a schema is a rooted contiguous tree fragment. These schemata divide the space of programs into subspaces containing programs of different sizes and shapes. The order $o(H)$ is the number of defining symbols (non-$\#$ nodes) $H$ contains. This definition uses positional information and does not allow multiple copies of one schema in a given program. This makes the calculations needed for a schema theorem much nicer but also restricts what is meant by a ``good" block of genetic material. The pros and cons will be discussed more in the conclusion. His work yields this schema theorem:

$$E(m(H, t + 1)) \geq m(H,t)\frac{f(H, t)}{\overline{f}(t)} \cdot \left(1-(p_m+p_c) \cdot \sum_{h\in H\cap Pop(t)} \frac{o(H)f(h)}{N(h)\cdot \sum_{h\in H\cap Pop(t)}f(h)}\right)$$

Here $N(h)$ is the size of a program $h$ matching $H$.

\subsection{Poli and Langdon's Definition}

Poli and Langdon similarly defined schema using positional data to avoid the problem of multiple instances of a schema appearing in the same individual. We highlight their definition, as it will be the one used for the remainder of the paper:

\begin{defn} A \textit{genetic program schema} is a rooted tree composed of nodes from the set $F \cup T \cup \{=\}$ where $F$ and $T$ are the function and terminal sets, and $=$ is a ``don't care" symbol which can be any arity needed. 
\end{defn}

With this definition, schemata partition the program space into subspaces of programs of fixed size and shape. Furthermore, the effect of genetic operators on schemata are now much easier to evaluate. 

\begin{notation}\label{GPnotation+}

For a fixed genetic programming schema $H$:

\begin{enumerate}
\item $m(H,t)$ is the number of instances of $H$ at generation $t$.

\item The order $o(H)$ is the number of non-$=$ symbols.

\item The length $N(H)$ is the total number of nodes in the schema

\item The defining length $L(H)$ is the number of links in the minimum tree fragment containing all non-$=$ symbols

\item $N(H)$ is the number of nodes of the individual

\item $p_d(t)$ is the fragility of $H$ in generation $t$

\item $G(H)$ be the zeroth order schema with the same structure as $H$ but all nodes being $=$ symbols

\item $P_{d_c}(H,h,t)$ is the probability of disruption due to crossover of schema $H$ in program $h$ in generation $t$. Taking the average over all $h$ yields $\overline{P}_{d_c}(H,t)$. Analogously we obtain $P_{d_m}$ for mutation.

\item The notation for $f$ and $\overline{f}$ is the same as in Notation \ref{notation}.

\end{enumerate}

\end{notation}

These quantities are independent of the shape and size of the programs. This definition is lower level than Rosca's definition above in that a smaller number of trees can be represented by schemata. Still, the trees represented by other schema can be represented by collections of genetic programming schemata and the converse is not true. Using this notation, we obtain:

\begin{theorem}[Genetic Programming Schema Theorem] \label{thm:GP} In a generational genetic program with fitness proportional selection, one-point crossover, and point mutation,

$$E(m(H,t+1))\geq m(H,t)\frac{f(H,t)}{\overline{f}(t)}(1-p_m)^{o(H)} \times$$

$$\left(1-p_c\left(p_d(t)\left(1-\frac{m(G(H),t)f(G(H),t)}{n\cdot \overline{f}(t)}\right)+\frac{d(H)m(G(H),t)f(G(H),t)-m(H,t)f(H,t))}{(N(H)-1)(n\cdot \overline{f}(t)}\right)\right)$$

\end{theorem}

Point mutation works by substituting a function node with another of the same arity or a terminal with another terminal. One-point crossover is defined as follows. First, identify the parts of both trees with the same topology (i.e. same arity nodes in the same places). Call this set of nodes the \textit{common region}. Then select a random crossover point and create the offspring with the part of parent 2 from below the crossover point and the rest from parent 1 above the crossover point. For the purposes of schema theory this crossover is better than standard genetic programming crossover (where a point is chosen randomly in each parent and crossover does not worry about the topology).

Note that one-point crossover yields children whose depth is no greater than that of the parents. This has the potential to help prevent bloat, though the author does not know whether this idea was ever explored. In (Poli and Langdon 1998) the authors also defined uniform crossover where ``the offspring is created by independently swapping the nodes in the common region with a uniform probability." For function nodes on the boundary of the common region, the nodes below them are also swapped. These two crossovers were motivated completely by schema theory but have become popular in their own right to practitioners since their creation. The problem of genetic programming for standard crossover was too hard to solve at this time, and was not solved till (Poli 2003a) and (Poli 2003b), which gave a Genetic Programming Schema Theorem for all homologous crossovers. The notion of a common region was not truly formalized until (Poli 2003a).

The Genetic Programming Schema Theorem is proven using basic probability theory applied to the four cases depending on whether the parents are both in $G(H)$, both not in $G(H)$, or only one is in $G(H)$. Analysis of the Genetic Programming Schema Theorem shows that the probability of schema disruption is very high at the beginning of a run. Furthermore, diversity of shapes and sizes will decrease over time and so a schema $H$ with above average fitness and short defining length will tend to survive better than other schemata. 

If all programs have the same structure then a genetic program with one-point crossover is nothing more than a genetic algorithm. In this situation, the theory of genetic programming limits to the theory of genetic algorithms as $t \to \infty$. Perhaps most important is the observation that two competitions are occurring in genetic programming. First, there is competition among programs with different $G(H)$. Then once only a few such hyperspaces are left there is competition within those hyperspaces on the basis of defining length. It is in the second phase that a genetic program acts like a genetic algorithm.

The definition of a genetic program schema was generalized twice more. The first generalization occurred in (Poli, 2000c) with the notion of a hyperschema.

\begin{defn} A \textit{hyperschema} is a rooted tree composed of nodes from the set $F \cup T \cup \{=,\#\}$ where $F$ and $T$ are the function and terminal sets, the symbol $=$ means ``don't care" for exactly one node, and $\#$ means any valid subtree. The special symbols can take any arity needed. 
\end{defn}

This definition generalizes both the definition of a genetic program schema and Rosca's definition. Using this definition, Poli generalized all his versions of the schema theorem (addressed below) and obtained cleaner statements, tighter bounds, and better proofs. Additionally, Poli used these results to argue for the existence of building blocks in genetic programming.

The final generalization was in (Poli, 2003b) with the notion of a variable arity hyperschema. There are also versions of schema theory in this context, and we leave it to the interested reader to investigate.

\begin{defn}
A \textit{variable arity hyperschema} is a rooted tree composed of internal nodes from the set $F \cup \{=,\#\}$ and leaves from $T \cup \{=,\#\}$ where the symbol $=$ means ``don't care" for exactly one node, the leaf $\#$ stands for any valid subtree, and the function $\#$ stands for a function with unspecified arguments which can be any valid subtree and arity $\geq$ the number of subtrees connected to it.
\end{defn}

In (Poli and Langdon, 1998) extensive experiments were performed on a genetic program to record all schema in the population, their average fitnesses, the population average fitness, the number of programs sampling a schema, the length, order, and defining length of schema, and schema diversity. The results confirmed a several conjectures. First, schema disruption is frequent at the beginning of a run before crossover counteracts the effects of selection. If selection only is acting, then schema with below average fitness disappear. Second, without mutation the population will start converging quickly and short schemata with above average fitness will have low disruption probability. Third, the average deviation in fitness of high-order schemata is larger than that for low-order schemata.

More surprisingly, this experiment points to the conclusion that building blocks do not grow exponentially. Rather, the growth function was not at all monotonic. The authors suggest that genetic drift and a small population size may have led to this conclusion. They suggested that to obtain achieve a better understanding of the genetic programming building blocks it may be necessary to study the propogation of structures which are functionally equivalent as programs but which are structurally different (now called ``phenotypical schemata").

It was found that initially, one-point crossover is as disruptive as standard genetic programming crossover, but as the run progresses it becomes much less disruptive. The disruption probability varies greatly from one generation to the next and therefore should be considered as a random variable (necessitating the conditional schema theorems of Section \ref{sec:conditional}). Finally, the genetic program does indeed asymptotically tend to a genetic algorithm, so nothing is lost by working in the setting of genetic programs for the rest of the paper. 

\section{Exact Schema Theory} \label{sec:exact}

Exact schema theory was originally created by Stevens and Waelbroeck in 1999. Poli expanded their work to create an exact version of the Genetic Programming Schema Theorem. The goal of exact schema theory is to obtain an equality rather than an inequality in the Schema Theorem by factoring the constructive forces of crossover (Poli 2001a). The Exact Schema Theorem applies to both genetic programming and variable length genetic algorithms and so it answers the first major criticism of schema theory.

Before continuing, let us fix some terms. A \textit{microscopic quantity} will mean a property of single strings/programs. A \textit{macroscopic quantity} will mean a property of a larger set of individuals such as average fitness. Resolving the criticisms of schema theory required Poli to produce tighter bounds on the expected values in the various schema theorems. This was his motivation for considering microscopic quantities. One of Poli's major focuses during this time (e.g. Poli 2001a; Poli 2003b) was expressing schema theorems in terms of purely microscopic quantities and then translating these statements so they involve only macroscopic quantities. This had the desired effect of producing exact schema theorems and also provided useful tools for future researchers interested in this area. The ability to express expected properties of a schema $H$ in terms of only properties of lower order schemata will be valuable if the connection between schema and genetic program efficacy is ever to be fully understood. 

The macroscopic versions of schema theorems tend to be more approximate but also much simpler and easier to analyze. The microscopic versions are more exact but there are many more parameters and degrees of freedom. The resulting exact models are typically huge and hard to manage computationally but more mathematically sound. The key to the exact schema theory is comparing genetic algorithm models based on whether they are approximate or exact, whether the predicted quantities are microscopic or macroscopic, and whether the predicting quantities are microscopic or macroscopic. As always, we pause to introduce notation, building upon Notation \ref{notation}:

\begin{notation} \label{exact-GA-notation} For a genetic algorithm schema $H$ in a population of strings of length $\ell$:

\begin{enumerate}
\item Let $\alpha(H, t)$ be the probability that a newly created individual will sample $H$. Call this the total transmission probability for $H$ in generation $t$. 

\item Let $p(H, t)$ be the selection probability of $H$ at generation $t$

\item Let $L(H, i)$ be the schema obtained from $H$ by replacing all elements from position $i+1$ to position $\ell$ with $\ast$

\item Let $R(H, i)$ be the schema obtained from $H$ by replacing all elements from position $\ell$ to position $i$ with $\ast$.

\end{enumerate}
\end{notation}

The reason for considering the truncations $L(H,i)$ and $R(H,i)$ become clear in the statement of the theorem:

\begin{theorem}[Exact Schema Theorem for Genetic Algorithms] In a population of size $n$, $E(m(H, t + 1)) = n \alpha(H, t)$ where

$$\alpha(H,t) = (1-p_c)p(H,t)+\frac{p_c}{\ell}\sum_{i=0}^{\ell-1}p(L(H,i),t)p(R(H,i),t)$$

\end{theorem}

Moving now to genetic programming schema, we introduce the necessary notation (building on Notation \ref{GPnotation+}):

\begin{notation} \label{exact-GP-notation} For a genetic program $j$:

\begin{enumerate}

\item Let $p_j^d$ be the probability that crossover in an active block of $j$ decreases fitness

\item Let $C_j^a$ equal the number of nodes in $j$

\item Let $C^e_j$ equal the number of nodes in the active part of $j$

\item Let $f_j$ denote the fitness of $j$

\item Define the \textit{effective  fitness} of $j$ to be $f_j^e = f_j \cdot \left(1-p_c\cdot \frac{C_j^e}{C_j^a}\cdot p_j^d\right)$

\item Let $P_j^t$ be the proportion of programs $j$ at generation $t$. It can be shown that  $P_j^{t+1} \approx P_j^t \cdot f_j^e/\overline{f}(t)$

\end{enumerate}
\end{notation}

Effective fitness formalizes a concept used by P. Nordin and W. Banzhaf in two papers to explain the reason for bloat and active-code compression. The result about $P_j^{t+1}$ describes ``the proliferation of individuals from one generation to the next." The effective fitness of $j$ is an approximation to the effective fitness of a genetic program individual. The effective fitness of a schema $H$ is defined by

$$f_e(H, t) = \frac{\alpha(H,t)}{p(H, t)} f(H, t) = f(H, t) \left(1- p_c\cdot \sum_{i\in B(H)}\left(1-\frac{p(L(H,i),t)p(R(H,i),t)}{p(H,t)}\right)\right)$$

where $B(H)$ is the set of crossover points within the defining length. Another useful concept is the operator adjusted fitness $f_a(H, t) = f(H, t)(1-p_c\frac{d(H)}{\ell-1}-p_mo(H))$.  This has been used to give a simplified version of the original Schema Theorem. It can also be used to formally define the notion of a deceptive problem. A \textit{deceptive problem} is one for which the optima of $f_a$ does not match the optima of $f$. Another way to define deception is using channels for creating instances of $H$. A channel is deceptive if $p(L(H, t), t)p(L(R, t), t) < p(H, t)$.

In (Poli and McPhee, 2001a) genetic programming schema theory is generalized to subtree mutation and headless chicken crossover. In 2003, Poli and McPhee wrote a 2-paper series which generalized this work and gave a General Schema Theorem for Genetic Programming which was exact, brought in conditional effects, could be formulated in terms of both microscopic and macroscopic terms, and held for variable-arity hyperschema.

\begin{theorem}[General Schema Theorem]
Using Notation \ref{exact-GA-notation}, there is a function $a_c(H,t)$ such that

$$a(H,t) = (1-p_c)p(H,t) + p_c a_c(H,t)$$
\end{theorem}

With this theorem, all previous schema theorems can be seen as computations of $a_c(H,t)$ under the various choices for selection, mutation, and crossover. The 2-paper series of 2003 culminated in the Microscopic and Macroscopic Exact Schema Theorems, which generalized all previous papers. These theorems work for any homologous crossover and virtually all subtree-swapping crossovers including standard genetic programming crossover, one-point crossover, context-preserving crossover, size-fair crossover, uniform crossover, and strongly-typed genetic programming crossover. 

We will now state these general theorems, but first we require the necessary notation. In order to mimic the truncations of Notation \ref{exact-GA-notation} for a hyperschema, we need to replace $L(H, t)$ and $R(H, t)$ by $u(H, i)$ and $l(H, i)$ which are the schemata obtained respectively by replacing all the nodes below $i$ with an $=$ and replacing all nodes not below $i$ with $=$. We generalize these to the \textit{upper building block hyperschema} $U(H, i)$ where we replace the subtree below $i$ with $\#$, and the \textit{lower building block hyperschema} $L(H, i)$ where we replace all nodes between the root and $i$ with an $=$ symbol. If $i$ is in the common region then these are empty sets.

\begin{notation} \label{exact-hyperschema-notation}

For a fixed hyperschema $H$:

\begin{enumerate}
\item Let $\alpha(H, t)$ be the probability that a newly created individual will sample $H$.
\item Let $U(H,i)$ denote the upper building block hyperschema, and let $L(H,i)$ denote the lower building block hyperschema introduced above
\item Let $NC(h_1, h_2)$ be the number of nodes in the common region between $h_1$ and $h_2$
\item Let $C(h_1, h_2)$ be the set of indices of common region crossover points, 
\item Let $\delta(x)$ be the Kronecker-Delta function (1 if $x$ is true and 0 otherwise)
\end{enumerate}
\end{notation}

Using this notation, we obtain:

\begin{theorem}[Microscopic Exact Genetic Programming Schema Theorem] For fixed size and shape genetic program schema $H$ under 1-point crossover and no mutation,

$$\alpha(H,t)=(1-p_c)p(H,t) + p_c \cdot \sum_{h_1,h_2\in Pop(t)}\frac{p(h_1,t)p(h_2,t)}{NC(h_1,h_2)} \cdot \sum_{i\in C(h_1,h_2)} \delta(h_1\in L(H,i))\delta(h_2\in U(H,i))$$
\end{theorem}

\begin{theorem}[Genetic Programming Schema Theorem with Schema Creation Correction] For a fixed size and shape genetic program schema $H$ under 1-point crossover and no mutation,

$$\alpha(H,t)\geq (1-p_c)p(H,t) + \frac{p_c}{\ell} \sum_{i=0}^{\ell-1}p(L(H,i)\cap G(H),t)p(U(H,i)\cap G(H),t)$$

with equality when all the programs are in $G(H)$
\end{theorem}

The difference between these two theorems is denoted $\Delta \alpha(H, t)$ and is sometimes called the \textit{Schema Creation Correction}. This is because the second theorem provides a better estimate if $p_d(t) = 1$. To obtain the final macroscopic theorem, which generalizes all previous schema theorems for genetic programming and genetic algorithms, label all possible schema shapes $G_1, G_2,\dots$. Carefully taking considerations related to shape into account, one can derive the following from the Microscopic Theorem:

\begin{theorem}[Macroscopic Exact Genetic Programming Schema Theorem] For fixed size and shape genetic program schema $H$ under 1-point crossover and no mutation,

$$\alpha(H,t)=(1-p_c)p(H,t) + p_c\cdot \sum_{j,k}\frac{1}{NC(G_j,G_k)} \cdot \sum_{i\in C(G_j,G_k)} p(L(H,i)\cap G_j,t)p(U(H,i)\cap G_k,t)$$

\end{theorem}

Note that all the above theorems were simplified and given easier proofs in (Poli, 2003b). As a corollary of the macroscopic theorem, we obtain the effective fitness for a genetic program as

$$f_e(H,t) = f(H,t)\left(1-p_c \left(1-\sum_{j,k} \sum_{i\in C(G_j,G_k)} 
\frac{p(L(H,i)\cap G(H),t)p(U(H,i)\cap G(H),t)}{NC(G_j,G_k)p(H,t)}\right)\right)$$

This fact has many applications which are discussed in Section \ref{sec:applications}. 

\section{Schema Theorems without the Expectation Operator} \label{sec:expectation}

As mentioned in the introduction, one criticism of schema theory was its reliance upon the expectation operator. While mathematicians may find this criticism difficult to comprehend, since the expectation operator is so ubiquitous, Poli responded to this criticism by creating a version of the Schema Theorem without the expectation operator (Poli, 1999). His method was to use Chebychev's Inequality from probability theory, which states:

$$P(|X -\mu| < k\sigma) \geq 1-1/k^2 \mbox{ for k any constant}$$

Abusing notation, let $\alpha(H,t)$ now denote the probability that $H$ survives \textit{or is created} after variation in generation $t$. As $\alpha$ forms a binomial distribution we have $\mu = n\alpha$ and $\sigma^2 = n\alpha (1-\alpha )$. Chebychev's Inequality gives:

\begin{theorem}[Two-sided probabilistic Schema Theorem] $$P \left( |m(H, t + 1) - n\alpha| \leq k\sqrt{n\alpha (1-\alpha)}\right) \geq 1-\frac{1}{k^2}$$
\end{theorem}

\begin{theorem}[Probabilistic Schema Theorem] $P \left(m(H, t + 1) > n\alpha - k \sqrt{n \alpha(1-\alpha)}\right) \geq 1-\frac{1}{k^2}$
\end{theorem}

In (Poli, 2000c) and (Poli, Langdon, and O'Reilly, 1998), Poli asked whether the lower bound in the Schema Theorem was reliable. This led him to investigate the impact of variance on schema transmission. Let $p_s(H, t)$ be the probability that individuals in $H$ will survive crossover and let $p_c(H, t)$ be the probability that offspring sampling $H$ will be created by crossover between parents not sampling $H$. Poli observed that this selection/crossover process is a Bernoulli trial and thus we are dealing with a binomial stochastic variable:

$$Pr(m(H, t + 1) = k) =  {n\choose k} \alpha(H, t)^k(1-\alpha(H, t))^{n-k}$$

$$E(m(H, t + 1)) = n\alpha(H, t) \mbox{ and }Var(m(H, t + 1)) = n\alpha(H, t)(1-\alpha(H, t))$$

These equations immediately yield an improvement to the Probabilistic Schema Theorem:

\begin{theorem} \label{thm:prob}
$Pr(m(H, t + 1)\geq x) = \sum_{k=x}^n {n\choose k} \alpha(H,t)^k (1-\alpha(H,t))^{n-k}$
\end{theorem}

This theorem holds regardless of the representation adopted, operators used, or definition of schema. Thus, it applies to genetic algorithms with bit-string representations of varying length. This is one of several instances where genetic programming schema theory yields useful applications to genetic algorithm schema theory and vice versa. From this theorem we go in two directions. First, we define the \textit{signal-to-noise ratio} 

$$\left(\frac{S}{N}\right) = \frac{E(m(H,t+1))}{\sqrt{Var(m(H,t+1))}} = \sqrt{n}\sqrt{\frac{\alpha(H,t)}{1-\alpha(H,t)}}$$

When this number is large the number of schemata in generation $t + 1$ will be very close to the expected number. When the ratio is small the number of schemata in generation $t+1$ will be effectively random. Poli shows that as $\alpha (H, t) \to 1$ the ratio approaches infinity. A corollary of these calculations is the probability of extinction in any generation:

$$Pr(m(H,t+1)=0) = (1-\alpha(H,t))^n = e^{n\log(1-\alpha(H,t))} \leq \frac{1}{\left(\frac{S}{N}\right)^2}$$

Clearly this quantity approaches zero as $\alpha(H, t)$ grows and so heuristically a schema is expected to survive if $\alpha(H, t) > 4/n$. However, newly created schema are very likely to go extinct. Calculations show the probability of extinction in the first generation after creation is above $37\%$. The probability of extinction within the first two generations is above $50\%$. This evidence suggests that more selection pressure or a steady-state genetic algorithm will be more likely to preserve high fitness newly created schemata. These heuristics come from useful bounds developed on the expectation and variance of $m(H, t +1)$. These bounds can be found in (Poli, Langdon, and O'Reilly, 1998). Combining the probabilistic theorems with the exact, microscopic, and macroscopic theorems gives a host of new and powerful theorems. Many of these theorems have not been explicitly stated, and this is one area for future research discussed in Section \ref{sec:future}.

The second direction to go from the probabilistic schema theorem is to fix the value of $x$ and attempt to solve for the right-hand side $y$ of the equation. This is what Poli refers to as ``predicting the past from the future" because we are fixing the future and finding what we need in the past to guarantee that future. Unfortunately, the solution is expressed in terms of the hypergeometric probability distribution (see Poli, 1999), so this is a hard problem. Some of the necessary mathematics has been done, and the $\tilde{\alpha}$ from the next section is one such inverse function. An application of Chebychev's Inequality using $k = (1-y)^{-1/2}$ can give a simpler answer but one without as much accuracy. At this point Poli also considered the effect of using Chernoff-Hoeffding bounds rather than the one-sided Chebychev inequality. The result is a slightly better bound on $y$ from $x$ and also confidence intervals. A more careful application of the various inequalities in probability theory may yield yet tighter bounds.

\section{Conditional Schema Theorems} \label{sec:conditional}

In 1997, 1998, and 1999 Fogel and Ghozeli published three papers claiming the Schema Theorem fails in the presence of a noisy or stochastic fitness function. They correctly pointed out an important bias in the sampling of schemata which comes from the use of fitness proportional selection in the presence of stochastic effects. Their key point was that $E \left(\frac{f(H,t)}{f(H,t)+f(H',t)}\right)$ must be used to calculate the correct proportion of individuals sampling $H$ when $H'$ is a competing schema.

In response, Poli reinterpreted the Schema Theorem as a conditional statement about random variables (Poli 2000a; Poli 2000b). In this form, the theorem now takes the form. In generation $t$, there is a constant $a \in [0,1]$ which contains information pertaining to the conditional expected value of $m(H,t+1)$, and the following equality is satisfied:

$$E(m(H, t + 1)| \alpha(H, t)=a)=na \mbox{ where n the is population size}$$

In order for this theorem to be useful, bounds on $a$ must be formulated. This is done below. First, the definition of expectation yields 

$$E(m(H, t + 1)) = \int_0^1 E(m(H, t + 1)|\alpha(H, t) = a)pdf(a)da$$

where $pdf(a)$ is the probability density function of $\alpha(H, t)$. Similarly, the selection-only Schema Theorem can be translated into a conditional statement. Let $f(H,t)$ denote the fitness of $H$ and $m(H,t)$ the number of instances of $H$ in generation $t$. Then the expected number of instances in the generation $t+1$ in a selection-only model is $m(H,t)\cdot f(H,t)/\overline{f}(t)$. Bringing in the effects of crossover and mutation yields the following formula, which makes no assumptions on the independence of the random variables involved:

$$E\left(\frac{m(H,t+1)}{n}\right) = E(\alpha(H,t))$$

This result can be specialized to the case of two competing schemata by integrating the conditional expectation function times the joint probability density function of the two schemata with respect to both fitness functions. Explicit formulas provided in (Poli, 2000b) serve the purpose required by Fogel and Ghozeli. Furthermore, this theorem can be used to predict the expected fitness distribution in generation $t+1$ if it is known in generation $t$. The considerations in this paper culminated in the following two theorems, taken from (Poli, 2000a):

\begin{theorem}[Conditional Probabilistic Schema Theorem] Let $E$ be the event that the following inequality is satisfied:

$$(1-p_c)\frac{m(H,t)f(H,t)}{n\overline{f}(t)} + \frac{p_c}{(\ell-1)n^2\overline{f}^2(t)} \cdot \sum_{i=1}^{\ell-1}m(L(H,i),t)f(L(H,i),t)m(R(H,i),t)f(R(H,i),t) \geq \tilde{\alpha}(k,x,n))$$

$$\mbox{ where } \tilde{\alpha}(k,x,n)) = \frac{n(k^2+2x)+k\sqrt{n^2k^2+4nx(n-x)}}{2n(k^2+n)}$$

Then the probability that $m(H,t+1)>x$ given that $E$ occurs is at least $1-\frac{1}{k^2}$, i.e. $Pr(m(H,t+1)>x | E) \geq 1-\frac{1}{k^2}$

\end{theorem}

Here $\tilde{\alpha}$ is an inverse function obtained by solving Theorem \ref{thm:prob} for $x$. It is a continuous increasing function of $x$. 

This theorem is proven using facts about continuous, differentiable maps with positive second derivative. With this theorem, Poli was able to obtain the Conditional Recursive Schema Theorem. For simplicity, let $p_c = 1$, assume the schema fitnesses and population fitnesses are known, let $P$ be the probability that the above equation is satisfied, and use Notation \ref{exact-GA-notation}. If $X$ is any random variable let $\cyclic{X}$ be any particular explicit value of $X$.

\begin{theorem}[Conditional Recursive Schema Theorem] \label{thm:recursive} For any choice of constants $M_H,M_L,M_R \in [0,n]$ and $i\in \{1,\dots,\ell-1\}$, consider the events 

$$\mu_i = \left\{ M_LM_R > \frac{\tilde{\alpha}(k,M_H,n)(\ell-1)n^2\cyclic{\overline{f}(t)}^2}{\cyclic{f(L(H,i),t)}\cyclic{f(R(H,i),t)}}\right\}$$

$$\mbox{ and } \phi_i = \{\overline{f}(t) = \cyclic{\overline{f}(t)}, f(L(H,i),t)=\cyclic{f(L(H,i),t)}, f(R(H,i),t)=\cyclic{f(R(H,i),t)}\}$$

Then the probability that $m(H,t+1)>M_H$ given $\mu_i$ and $\phi_i$ is at least

$$\left(1-\frac{1}{k^{2}}\right)\cdot [Pr(m(L(H,i),t) > M_L | \mu_i,\phi_i)+Pr(m(R,H,i),t) > M_R | \mu_i,\phi_i)-1]$$
\end{theorem}

In the proof, $\mu_i$ serves to guarantee $M_L$ and $M_R$ are appropriate, while $\phi_i$ restricts the number of cases needed in the proof to just two. This theorem answers completely Fogel and Ghozeli's criticism as well as the second major criticism from the mid 1990s. This theorem can be applied to the events on the right-hand side recursively and thus it gives a way to list the conditions necessary on the initial population to ensure convergence, again assuming one knows the fitnesses and building blocks.

A nice application of this theorem is that the lower bound for the probability of convergence in generation $t$ is a linear combination of probabilities that there are enough building blocks in the initial population. This makes clear the relationship between population size, schema fitness, and convergence probability. Maximizing the lower bound on the probability of convergence provides a lower bound on population size, and thereby helps programmers determine an appropriate value for the population size parameter.

\section{Relationship between Schema Theory and Markov Models} \label{sec:markov}

Schema theories up until 2001 were purely macroscopic models of genetic algorithms. In her 1996 textbook, Mitchell claims that models based on Markov chains and statistical mechanics will be necessary to formalize the theory of genetic algorithms. Markov chain models are generally exact and fully microscopic, so they hold more appeal for computer scientists looking for accuracy. The cost is that Markov models have many degrees of freedom and are difficult to derive and use. For examples of such models, see (Davis and Principe, 1993; Nix and Vose, 1992). Statistical mechanics models are macroscopic so they are simpler but also less accurate. For examples of such models see (Pr\"{u}gel-Bennett and Shapiro, 1994). Markov models have also been applied to genetic programming (Mitavskiy and Rowe, 2006; Poli et al., 2004; Poli, Rowe, and McPhee, 2001).

For most of the 1990s macroscopic models dominated the field of genetic algorithm modeling. (Poli, 2001a) helped combat this by providing an exact microscopic model, but there are still many more macroscopic models than microscopic. Aggregation of states in Markov chain models provide another way to move from the microscopic to the macroscopic. Again, the benefit of such a move is in simpler equations, but the downside is a loss of accuracy. Very few of these models explicitly factor in the fitness function, so the fitness landscape cannot be seen even if we get a the program space is well understood. This is one reason why genetic programming schema theory is useful: it provides a new point of view regarding this failure of Markov chain models. Exact genetic programming models are designed to explicitly determine how selection and the variation operators affect sample program space.

While the 1990s saw a competition between Markov models and schema theory, both have provided useful applications and theory, and the two now seem to coexist as alternative approaches. These two approaches are linked in (Poli-McPhee-Rowe, 2004). In this paper, the authors successfully applied schema theory to determine the biases of the variation operators (see also Poli and McPhee, 2002) and helped get a better hold on the program space. Furthermore, the authors create a new Markov chain model and use it to analyze in detail $0/1$ trees.

This work required the use of crossover masks, which generalize all possible choices of crossover operator. For fixed-length binary strings, a \textit{crossover mask} is a binary string which tells how crossover is done. In this string, a 1 is an instruction to choose the allele at this position from parent $A$ while a 0 is an instruction to take the allele from parent $B$. For each mask $i$, let $p_i$ be the probability this mask is selected. The distribution of the $p_i$ is the recombination distribution. Crossover masks can also be generalized to genetic programming via trees with the same size and shape as the common region. Similarly, there is a generalized recombination distribution and a building block generating function for genetic programs. Denote this function by $\Gamma(H, i)$. It returns the empty set if $i$ contains a node not in $H$ and otherwise it returns the hyperschema with size and shape matching $H$ but with $=$ nodes everywhere. Letting $\overline{i}$ be the tree complementary to $i$, it turns out that $\Gamma(H, i)$ and $\Gamma(H, \overline{i})$ generalize $L(H, i)$ and $U(H, i)$ from Notation \ref{exact-GA-notation}.

The following lemma is used to prove both microscopic and macroscopic exact theorems stating the total transmission probability for a genetic program schema using homologous crossover. This is the heart of the 2004 paper.

\begin{lemma} If $P_1 \in \Gamma(H, i)$ and $P_2\in \Gamma(H, \overline{i})$ then crossing $P_1$ and $P_2$ using mask $i$ gives an offspring matching $H$. Conversely, if crossover according to $i$ yields an individual matching $H$ then one of the parents must have come from $\Gamma(H, \overline{i})$ and the other from $\Gamma(H, i)$.
\end{lemma}

Both Markov chain models and schema theory are attempts to look carefully at the generation-to-generation behavior of a genetic algorithm or genetic program. In the conclusion of (Poli-McPhee-Rowe, 2004), the authors claim that exact schema models are simply different representations of Markov chain models. If this is true then these two fields are equivalent, and there are many areas where results from one may be applied to the other. The authors support this claim by noting the use of dynamic building blocks in the formulation of exact schema theory. The authors leave open the details of making this connection precise, so we have included this as an open problem in Section \ref{sec:future}. 

\section{Applications of Schema Theory} \label{sec:applications}

In general, theorists in this field wish to understand the benefits of schema theory over other theories which attempt to explain the efficacy of genetic algorithms. Over a number of years, Poli laid the groundwork for schema theory and resolved the criticisms which this field faced. Practitioners tend to be more interested in choosing the correct representation of a problem, choice of operators, choice of fitness function, settings of parameters, size of population, number of runs, etc. The problem of combating bloat is also important. There have also been applications to fields other than genetic programming, and these are discussed at the end of the section.

\subsection{Applications to Genetic Programs}

The considerations in Section \ref{sec:GP} provide comparisons between the various ways to represent a problem. Finding a relationship between the choice of representation and schema behavior can help practitioners choose the best representation for their chosen application. Another instance where schema theory helps practitioners was discussed at the end of Section \ref{sec:conditional}. Here schema theory provides a lower bound on population size and thereby helps practitioners determine a good value for the population size parameter.

Regarding the problem of bloat, (Poli, 2001a) shows how to use the macroscopic exact genetic programming schema theorem and its corollary to determine when there is an effective fitness advantage in having a large amount of inactive code. Thus, bloat can at times be necessary for the success of the genetic program. However, because bloat often slows down a genetic program, practitioners can also use the considerations in (Poli, 2001a) to avoid situations which lead to bloat. 

Schema theory has motivated new crossover functions (one-point and uniform) and the notion of a smooth operator, and these notions have been useful to practitioners. Furthermore, in (Poli and McPhee, 2002) different measurement functions are explored to investigate the behavior and biases of the variation operators and parameters. This led to new initialization strategies for genetic programming to optimize performance using knowledge of the variation operator biases. Finally, (Poli, 2001a) provides an exact formulation of which problems are hard for a genetic program to solve (see also Poli, 2003b).

To discuss other applications, we must first discuss the proof method in the 2003 paper of Poli and McPhee. This proof method relies on the \textit{Cartesian node reference system representation} of genetic programs as functions over the space $\mathbb{N}^2$ and the process of selection as a probability distribution over $\mathbb{N}^4$. This reference system consists of laying out the trees on a grid where each node has arity $a_m$. After doing so, there are $a_m^n$ nodes at depth $n$. Any node can be recovered from its depth $(d)$ and where it falls $(i)$ in that row. We list the relevant notation:

\begin{notation}\label{cartesian-notation}
\begin{enumerate}
\item The name function $N(d, i, h)$ returns the node in $h$ at position $(d, i)$. 

\item The size function $S(d, i, h)$ returns the number of nodes present in the subtree rooted at $(d, i)$. 

\item The arity function $A(d, i, h)$ returns the arity of the node at $(d, i)$. 

\item The type function $T (d, i, h)$ to return the data type of the node.

\item The function-node function $F (d, i, h)$ returns 1 if the node is a function and 0 otherwise.

\item The common region membership function $C(d, i, h, h)$ which returns 1 if $(d, i)$ is in the common region and 0 otherwise.

\item Define $p(d, i|h)$ as the probability that the $(d, i)$ node is selected in program $h$.

\item Define $p(d_1, i_1, d_2, i_2|h_1, h_2)$ as the probability that $(d_j, i_j)$ is selected in $h_j$.

\item Define $p(d_1, i_1, d_2, i_2) = p(d_1, i_1|h_1)p(d_2, i_2|h_2)$. A symmetric crossover is one for which $p(i, j|h_1, h_2) = p(j, i|h_2, h_1)$.

\end{enumerate}
\end{notation}

The microscopic and macroscopic schema theorems can be specialized for standard genetic programming crossover using this machinery. A corollary is the size-evolution equation for genetic programming with subtree-swapping crossover:

\begin{theorem}[Size Evolution Equation] Let $\mu$ be the mean size of a program in a genetic program population. If the genetic program uses a symmetric subtree-swapping crossover operator and no mutation then for fixed $(d, i)$ we have

$$E( \mu(t + 1)) = \sum_{h\in Pop(t)}S(h)p(h, t) =  \sum_k S(G_k)p(G_k, t)$$
\end{theorem}

The theorem tells us that the mean program size evolves as if selection alone were acting on the population. Thus, bloat is the effect of selective pressure and we can calculate the mean size of individuals at time $t$ in terms of the number $N$ of individuals sampling a schema $G_k$ and the proportion $\Phi$ of individuals of size and shape $G_k$: $E( \mu(t)) = \sum_k N(G_k)\Phi(G_k, t)$. This allows for the prediction and control of bloat, helping to solve a major open problem dating back to Koza's early work in genetic programming. In this application, controlling bloat can be achieved by acting on the selection probabilities to discourage growth, e.g. by creating holes in the fitness landscape which repel the population. Note that this equation was expanded and simplified by Poli and McPhee in 2008 to exactly formalize program size dynamics.

\begin{corollary} For a flat landscape, we have $E( \mu(t + 1)) = \mu(t)$
\end{corollary}

This has also been studied by Poli and McPhee for at landscapes with holes and spikes. This study can also be used to fine tune parameter settings for the variation operators and move towards optimal performance. In the same vein, (Poli et al, 2003) applied schema theory to look into bistability of a gene pool genetic algorithm with mutation. A bistable landscape is one with two stable fixed points on a single-peak fitness landscape. This paper led to a better overall understanding of mutation and also provided another example of an unexpected application of schema theory.

Another corollary of the size-evolution equation is the study of Crossover Bias Theory (Poli, 2008). This states that because crossover removes as much material as it adds (on average), crossover is not solely responsible for changes in mean program size. However, crossover does affect the higher moments of the distribution, pushing the population towards a Lagrange Distribution of the Second Kind. In this distribution smaller programs have significantly higher frequency than larger programs. Thus, larger programs have an evolutionary advantage over smaller programs and this forces the mean program size to increase over time.

\subsection{Applications to Other Fields}

After genetic programming schema theory provided a definition for the notion of a deceptive program (discussed in Section \ref{sec:exact}), (Langdon and Poli 1998) were able to investigate deception in the Ant problem. The Ant problem is a difficult search-space problem in which the goal is to program an artificial ant to follow the Santa Fe trail. A priori, this problem has nothing to do with schema theory, so this is a particularly striking application.

Another striking application is the relationship between schema theory and Markov models discussed at the end of Section \ref{sec:markov} and in (Poli-McPhee-Rowe, 2004). For instance, the authors hint that schema theory can be used to determine conditions under which the Markov model's transition matrix is ergodic. This relationship has the potential for many other future applications as well, some of which are discussed in Section \ref{sec:future}.

In (Poli and McPhee, 2001b), the authors specialized exact genetic programming schema theory to the case of linear structures of variable length (i.e. where the functions are all unary). For example, binary strings or programs with primitives of arity 1 only are linear structures. The authors then found a version of the theorem for standard crossover in the linear representation case. Finally, they considered fixed points of the schema equations which allowed them to prove that standard crossover has a bias which samples shorter structures exponentially more frequently than longer structures (Poli and McPhee, 2002). This led to a number of conclusions about linear systems (Poli, 2003b).

First, genetic programming crossover exerts a strong bias forcing the population towards a Gamma length distribution. This bias is strong enough to overpower the selection bias, so a practical application is to initialize a population so that the lengths begin by matching the Gamma distribution. Second, shorter than average structures are sampled exponentially more than longer ones, so the genetic program wastes time resampling. Thus, time could be saved by setting the mean length of the initial structures so that the sampling will occur where solutions are believed to be. Third, focusing on the distribution of primitives in the representation gives a generalization to Geiringer's Theorem and of the notion of linkage equilibrium to representations with non-fixed length (Poli, et al 2003b). It can be shown that the primitives in each individual tend to be swapped with those of other individuals and also to diffuse across the positions within the individual. This diffusive bias may also interfere with the selection bias and so should be avoided by initializing the population so that primitives are already uniformly distributed in each individual. Fourth, highly fit individuals may fail to transmit their genes, so this should be avoided by moving towards a steady-state model.

These four lessons show a very nice application of schema theory to a well-studied problem area. Linear structures enter in the following way. Given a genetic program made up of linear structures with only two functions and only two terminals, the concept of a genetic program schema corresponds exactly to that of a genetic algorithm schema. Similarly, 1-point genetic programming crossover corresponds exactly to 1-point genetic algorithm crossover. This gives yet another link between the theories of genetic programming and genetic algorithms.

\section{Future Directions} \label{sec:future}

As the huge number of papers listed above demonstrates, schema theory is certainly an interesting and rich field of study. Many early arguments based on schema theory demonstrated a lack of understanding and rigor on the part of those writing the papers, but this led to disapproval among computer scientists of the field as a whole. This disapproval chased away many who could have done work in schema theory, but thanks to the work of Riccardo Poli and his collaborators, schema theory survived this early setback and is becoming a popular field once more. Poli has demonstrated that all criticisms leveled at schema theory are either unfounded or can be fixed. He has created a huge number of schema theorems and related them to Markov models so that other computer scientists can use them effectively. Furthermore, Poli and others have found numerous surprising applications of schema theory. These include applications to the Ant problem, to controlling bloat, to the study of variation operators, to parameter setting, and to population initialization. There are also several extended examples such as linear structures where the insights from schema theory have been invaluable.

There are many more directions for development of schema theory. We begin with the easiest first. The next two problems build directly on the considerations discussed in the previous sections:

\begin{problem}
Extend the work discussed at the end of Section \ref{sec:expectation} and combine the Probabilistic Schema Theorems with the Exact, Microscopic, and Macroscopic Schema Theorems.
\end{problem}

\begin{problem}
Update the existing schema theorems to address the effects of mutation. Consider creating a mutation mask analogous to the crossover mask from Section \ref{sec:markov} so that all choices of mutation are handled at once.
\end{problem}

Once this is done, the natural next step would be to specialize the schema theorems for different mutation operators as has been done for different selection and crossover operators. This would likely lead to a better understanding of the role of mutation in genetic algorithms and genetic programming just as the schema theory above led to a better understanding of the role of crossover and selection. 

(Whitley, 1993) claims that mutation can be used as a hill-climbing mechanism and that genetic algorithms using only mutation can compete with genetic algorithms using crossover and a small mutation probability. Mutation often serves the role of searching locally, while crossover is more of a global search. It would be valuable to understand the relationship between local and global search, and the considerations within schema theory may help in obtaining this understanding.

A more difficult but more fruitful collection of problems is based on the considerations in Section \ref{sec:applications}.

\begin{problem}
Formulate schema theories for developmental genetic programming, Genetic Algorithm for Rule Set Production (GARP), learning classifier systems, and other systems commonly studied.
\end{problem}

Moving in a different direction, note that the development of schema theory has so far only made use of relatively elementary results from probability theory. Each time more advanced results were introduced they led to significant gains in schema theory. There are more advanced tools from probability theory which have not been used, and this provides a rich and fertile area for future work. 

For example, to the author's knowledge no one has tried to apply the study of martingales to the recursive conditional schema theorem. There is a well-developed study of conditional stochastic processes and many tools from this could be applied. It is possible that more advanced mathematical approaches would answer the remaining open problems in this field. Perhaps the biggest such problem is that schema theory and Markov models do not contain information about the fitness distribution in the search space, so we cannot hope to fully characterize genetic programming behavior (Poli, 2008).

\begin{problem}
Sharpen the inequalities in the various schema theorems by making use of more sophisticated tools from probability theory.
\end{problem}

\begin{problem}
Introduce the study of martingales into recursive conditional schema theory, and use this to obtain tighter bounds and to make better predictions for the behavior of evolutionary computation in the presence of conditional effects.
\end{problem}

\begin{problem} \label{prob:fitness}
Find a way to factor information about the fitness distribution into schema theory.
\end{problem}

In a related vein, there is still much work to be done towards relating schema theory to Markov models.

\begin{problem}
Fill in the details of the claims in the conclusion of (Poli-McPhee-Rowe, 2004) that exact schema models contain precisely the same information as Markov chain models.
\end{problem}

A reasonable place to begin working on this problem would be the remark in (Poli-McPhee-Rowe, 2004) that using their exact formulas for the probability that reproduction and recombination will create any specific program, ``a GP Markov chain model is then easily obtained by plugging this ingredient into a minor extension of Vose's model for genetic algorithms" (Poli-McPhee-Rowe, 2004). 

Once this has been done, results obtained by Vose and the Markov chain model can be generalized to apply to both genetic programming and variable-length genetic algorithms. For example, as pointed out in (Poli, 2003b), Markov chain models have the ability to calculate probability distributions over the space of all possible populations. If there is a version of this for schema theory then it might help to produce better bounds in schema theorems and to produce more accurate long term predictions.

\begin{problem}
Generalize Vose's results on the Markov chain model to apply to both genetic programming and variable-length genetic algorithms. 
\end{problem}

\begin{problem}
Use schema theory to determine conditions under which the Markov model's transition matrix is ergodic. 
\end{problem}

If Problem \ref{prob:fitness} can be solved and if the dictionary between schema theory and Markov models can be fleshed out, then this could also provide a way for Markov chain models to factor in information about the fitness distribution.

There are also hard foundational left to resolve in this field. The most famous was mentioned in the Introduction:

\begin{problem}
Use the theory of dynamical systems to formulate a precise mathematical statement of the building block hypothesis. Prove or disprove this statement.
\end{problem}

The next problem goes back to the comparison of definitions in Section \ref{sec:GAschema}. (Whigham, 1995) allowed multiple instances of a schema in a given tree and showed that this definition restricts to both the usual genetic programming schema theory and to the genetic algorithm schema theory. In biology, genetic material does not come in packages which corresponding to single ``good" templates. Rather, having multiple copies of a ``good" block of alleles or a ``good" template is rewarded. For artificial evolution to successfully move towards computational evolution it will be necessary to include the possibility of multiple instances of a schema. 

\begin{problem} Generalize the schema theorem so that it holds for genetic algorithm strings which allow multiple copies of a schema to be represented.
\end{problem}

Learning classifier systems provide one way in which to attack this problem. In particular, the use of numerosity in accuracy-based learning classifier systems like XCS could be adapted to create an initial model which contains the possibility of multiple schema instances. The schema theorem which works for variable-length genetic algorithms is a very nice step in this direction. 

The solution to these last problems may add another layer of complexity to the schema theorems, but with care the solution could also create a simpler overall theory. Furthermore, this theory would be robust enough to cope with self-modification mechanisms and feedback in artificial evolution. Generalizing schema theory in the direction of computational evolution would likely lead to good suggestions for ways to factor in this added layer of complexity. One crucial way to move towards computational evolution is by generalizing genetic programming schema theorems so they work when we don't restrict the size and shape of the program. Poli has avoided doing this because he believes ``simply getting to the point of stating exact models for these algorithms requires a lot of machinery" (Poli-McPhee-Rowe, 2004). But because there is so much machinery in probability theory which has not been exploited, it is not unreasonable to hope that generalizing schema theory will one day occur. More sophisticated mathematics should lead to a better grasp of the theory.

The application of genetic programming schema theory to genetic algorithm schema theory with a representation of variable-length bit strings (Poli, 2001a) will be very useful in the move from artificial evolution to computational evolution. This is because in biology the genome is not given by fixed length chromosomes. Rather, genetic material can come in all shapes and sizes. The existence of this genetic algorithm generalization should also give us all hope that it will be possible to generalize genetic programming schema theorems to work when the representation is not of a fixed length size and shape. The other applications of schema theory will also be useful in the move to computational evolution, and their existence suggests that schema theory will continue to give surprising applications in the future. If there is any lesson to be taken from Poli's work it is that schema theory is strong and can adapt to any problem thrown at it. Generalizing schema theory tends to make proofs easier to understand, so it's possible schema theory can lead the way in the move from artificial evolution to computational evolution.

\section{References}

[Altenberg, 1994] L. Altenberg, ``Emergent phenomena in genetic programming," in \textit{Evolutionary Programming - Proceedings of the Third Annual Conference} (A. V. Sebald and L. J. Fogel, eds.), pp. 233-241, World Scientific Publishing, 1994.

[Altenberg, 1995] L. Altenberg, ``The Schema Theorem and Price's Theorem," in \textit{Foundations of Genetic Algorithms 3} (L. D. Whitley and M. D. Vose, eds.), (Estes Park, Colorado, USA), pp. 23-49, Morgan Kaufmann, 31 July-2 Aug. 1995.

[Burjorjee, 2008] Burjorjee, K. M. ``The Fundamental Problem with the Building Block Hypothesis" submitted to the \textit{Evolutionary Computation Journal}, April 7, 2009.

[Davis and Principe, 1993] T. E. Davis and J. C. Principe. ``A Markov chain framework for the simple genetic algorithm," in \textit{Evolutionary Computation}, 1(3):269-288, 1993.

[Holland, 1992] Holland, J. (1992). \textit{Adaptation in Natural and Artificial Systems}. MIT Press, Cambridge, Massachusetts, second edition.

[Koza, 1992] Koza, J. R. (1992). \textit{Genetic Programming: On the Programming of Computers by Means of Natural Selection}. MIT Press.

[Langdon and Poli 1998] W.B. Langdon and R. Poli. ``Why Ants are Hard." Technical Report CSRP-98-4, University of Birmingham, School of Computer Science, January 1998.

[Mitavskiy and Rowe, 2006] B. Mitavskiy and J. Rowe. ``Some results about the markov chains associated to GPs and to general EAs." \textit{Theoretical Computer Science}, 361(1):72-110, 28 August 2006.

[Mitchell 1996] Mitchell, M. \textit{An Introduction to Genetic Algorithms}, MIT press, 1996.

[Nix and Vose, 1992] Nix, A. E., and Vose, M., D., ``Modeling Genetic Algorithms With Markov Chains," in \textit{Annals of Mathematics and Artificial Intelligence 5} (1992), pp. 79-88.

[Nix and Vose, 1992] A. E. Nix and M. D. Vose. ``Modeling genetic algorithms with Markov chains," in \textit{Annals of Mathematics and Artificial Intelligence}, 5:79-88, 1992.

[O'Reilly, 1995] O'Reilly, U.M. (1995). \textit{An Analysis of Genetic Programming}. PhD thesis, Carleton University, Ottawa-Carleton Institute for Computer Science, Ottawa, Ontario, Canada.

[O'Reilly and Oppacher, 1994] O'Reilly, U.M. and Oppacher, F. (1995). The troubling aspects of a building block hypothesis for genetic programming. In Whitley, L. D. and Vose, M. D., editors, \textit{Foundations of Genetic Algorithms 3}, pages 73-88, Estes Park, Colorado, USA. Morgan Kaufmann.

[Poli and Langdon, 1997] R. Poli and W. B. Langdon. A new schema theory for genetic programming with one-point crossover and point mutation. In J. R. Koza, K. Deb, M. Dorigo, D. B. Fogel, M. Garzon,H. Iba, and R. L. Riolo, editors, \textit{Genetic Programming 1997: Proceedings of the Second Annual Conference}, pages 278-285, Stanford University, CA, USA, 13-16 July 1997. Morgan Kaufmann.

[Poli and Langdon, 1998] R. Poli and W. B. Langdon, ``A review of theoretical and experimental results on schemata in genetic programming," in \textit{Proceedings of the First European Workshop on Genetic Programming} (W. Banzhaf, R. Poli, M. Schoenauer, and T. C. Fogarty, eds.), vol. 1391 of LNCS, (Paris), pp. 1-15, Springer-Verlag, 14-15 Apr. 1998.

[Poli, Langdon, O'Reilly 1998] R. Poli, W.B. Langdon, and U.-M. O'Reilly. ``Analysis of Schema Variance and short term extinction likelihoods," in \textit{Genetic Programming 1998: Proceedings of the Third Annual Conference}, pp. 284-292, Morgan Kaufmann, 22-25 July 1998.

[Poli 1999] Poli, R. Schema Theorems without Expectations. In Banzhaf, W., Daida, J., Eiben, A. E., Garzon, M. H., Honavar, V., Jakiela, M., and Smith, R.E., editors, \textit{Proceedings of the Genetic and Evolutionary Computing Confrence, volume 1}, page 806, Orlando, Florida, USA. Morgan Kaufmann.

[Poli 2000a] R. Poli. ``Recursive Conditional Schema Theorem, Convergence and Population Sizing in Genetic Algorithms," in \textit{Proceedings of the Foundations of Genetic Algorithms Workshop (FOGA 6)}, pages 146-163, 2000. Morgan Kaufmann.

[Poli 2000b] R. Poli. ``Why the schema theorem is correct also in the presence of stochastic effects," in \textit{Proceedings of the Congress on Evolutionary Computation (CEC 2000)}, pages 487{492, 2000.

[Poli 2000c] R. Poli, ``Hyperschema theory for GP with one-point crossover, building blocks, and some new results in GA theory," in \textit{Genetic Programming, Proceedings of EuroGP 2000} (R. Poli, W. Banzhaf, et al., eds.), Springer-Verlag, 15-16 Apr. 2000.

[Poli 2001a] R. Poli. ``Exact schema theory for genetic programming and variable-length genetic algorithms with one-point crossover," in \textit{Genetic Programming and Evolvable Machines}, 2(2): 123-163, 2001.

[Poli 2003a] R. Poli and N. F. McPhee. General schema theory for genetic programming with subtree swapping crossover: Part I. Evolutionary Computation, 11(1):53-66, March 2003a. URL http://cswww.essex.ac.uk/staff/rpoli/papers/ecj2003partI.pdf.

[Poli 2003b] R. Poli and N. F. McPhee. General schema theory for genetic programming with subtree swapping crossover: Part II. Evolutionary Computation, 11(2):169-206, June 2003b URL http://cswww.essex.ac.uk/staff/rpoli/papers/ecj2003partII.pdf.

[Poli-McPhee-Rowe, 2004] R. Poli, N. F. McPhee, and J. E. Rowe. Exact schema theory and Markov chain models for genetic programming and variable-length genetic algorithms with homologous crossover. Genetic Programming and Evolvable Machines, 5(1):31-70, March 2004. ISSN 1389-2576. URL: http://cswww.essex.ac.uk/staff/rpoli/papers/GPEM2004.pdf.

[Poli 2000d] R. Poli. ``Exact schema theorem and effective fitness for GP with one-point crossover." In D. Whitley, et al., editors, \textit{Proceedings of the Genetic and Evolutionary Computation Conference}, pages 469-476, Las Vegas, July 2000. Morgan Kaufmann.

[Poli, Rowe, and McPhee 2001] R. Poli, J. E. Rowe, and N. F. McPhee, ``Markov chain models for GP and variable-length GAs with homologous crossover," in \textit{Proceedings of the Genetic and Evolutionary Computation Conference (GECCO-2001)}, (San Francisco, California, USA), Morgan Kaufmann, 7-11 July 2001.

[Poli and McPhee 2001a] R. Poli and N. F. McPhee, ``Exact GP schema theory for headless chicken crossover and subtree mutation," in \textit{Proceedings of the 2001 Congress on Evolutionary Computation CEC 2001}, (Seoul, Korea), May 2001.

[Poli and McPhee 2001b] R. Poli and N. F. McPhee, ``Exact schema theorems for GP with one-point and standard crossover operating on linear structures and their application to the study of the evolution of size," in \textit{Genetic Programming, Proceedings of EuroGP 2001}, LNCS, (Milan), Springer-Verlag, 18-20 Apr. 2001.

[Poli 2001b] R. Poli. ``General schema theory for genetic programming with subtree-swapping crossover." In \textit{Genetic Programming, Proceedings of EuroGP 2001}, LNCS, Milan, 18-20 April 2001. Springer-Verlag.

[Poli and McPhee, 2002] N F McPhee and R Poli. ``Using schema theory to explore interactions of multiple operators" in \textit{GECCO 2002: Proceedings of the Genetic and Evolutionary Computation Conference} pp. 853{860. Morgan Kaufmann

[Poli 2003c] Poli, R. ``A simple but theoretically-motivated method to control bloat in genetic programming." In Ryan, C., Soule, T., Keijzer, M., Tsang, E., Poli, R., and Costa, E., editors, \textit{Genetic Programming, Proceedings of the 6th European Conference, EuroGP 2003}, LNCS, pages 211.223, Essex, UK. Springer-Verlag.

[Poli et al 2003] Alden H. Wright, Jonathan E. Rowe, Christopher R. Stephens and Riccardo Poli, ``Bistability in a Gene Pool GA with Mutation" in K. De Jong, R. Poli and J. Rowe, editors, \textit{Proceedings of the Foundations of Genetic Algorithm (FOGA-7) Workshop}, Torremolinos, Spain, 3-5 September 2002, Morgan Kaufmann, pages 63-80, 2003.

[Poli et al 2003b] R Poli, C R Stephens, A H Wright, and J E Rowe. ``A schema-theory-based extension of Geiringer's theorem for linear GP and variable-length GAs under homologous crossover" in \textit{Foundations of Genetic Algorithm 7 }(2003) pp. 45-62. Morgan Kaufmann

[Poli and Langdon 2006] R. Poli and W. B. Langdon. ``Efficient markov chain model of machine code program execution and halting." In R. L. Riolo, et al., editors, \textit{Genetic Programming Theory and Practice IV}, volume 5 of Genetic and Evolutionary Computation, chapter 13. Springer, Ann Arbor, 11-13 May 2006. ISBN 0-387-33375-4. URL http://www.cs.essex.ac.uk/sta /poli/papers/GPTP2006.pdf.

[Poli, 2008] R. Poli, W. B. Langdon, and N. F. McPhee. \textit{A field guide to genetic programming}. Published via http://lulu.com and freely available at http://www.gp- eld-guide.org.uk, 2008. (With contributions by J. R. Koza)

[Pr\"{u}gel-Bennett and Shapiro, 1994] A. Pr\"{u}gel-Bennett and J. L. Shapiro. An analysis of genetic algorithms using statistical mechanics. \textit{Physical Review Letters}, 72:1305-1309, 1994.

[Rosca 1997] J P Rosca. ``Analysis of complexity drift in genetic programming," in \textit{Genetic Programming 1997: Proceedings of the Second Annual Conference} pp. 286{294, 1997, Morgan Kaufmann.

[Rosca and Ballard 1999] J. P. Rosca and D. H. Ballard. ``Rooted-tree schemata in genetic programming." In L. Spector, et al., editors, \textit{Advances in Genetic Programming 3}, chapter 11, pages 243-271. MIT Press, Cambridge, MA, USA, June 1999. ISBN 0-262-19423-6.

[Stephens and Waelbroeck, 1999] C. R. Stephens and H. Waelbroeck, ``Schemata evolution and building blocks,"\textit{ Evolutionary Computation}, vol. 7, no. 2, pp. 109-124, 1999.

[Syswerda, 1989] Syswerda, G. ``Uniform crossover in genetic algorithms" in \textit{Proceedings of the Third International Conference on GA}, 1989, pp. 2-9. Morgan Kaufmann

[Whigham, 1995] Whigham, P. A. (1995). A schema theorem for context-free grammars. In \textit{1995 IEEE Conference on Evolutionary Computation, volume 1}, pages 178-181, Perth, Australia. IEEE Press.

[Whitley, 1993] Whitley, D. (1993). A genetic algorithm tutorial. Technical Report CS-93-103, Department of Computer Science, Colorado State University.

\end{document}